\documentclass[10pt,twocolumn]{article}

\usepackage{cvpr}
\usepackage{times}
\usepackage{epsfig}
\usepackage{graphicx}
\usepackage{amsmath,amsfonts,amsthm}
\usepackage{amssymb}
\usepackage[lined,ruled,commentsnumbered]{algorithm2e}
\usepackage{color} 
\usepackage{rotating}

\definecolor{P}{rgb}{0,0.6,0} 

\makeatletter
\renewcommand{\paragraph}{%
	\@startsection{paragraph}{4}%
	{\z@}{.2ex \@plus 1ex \@minus .1ex}{-1em}%
	{\normalfont\normalsize\bfseries}%
}
\makeatother

\usepackage{color}
\usepackage{multirow}
\usepackage[percent]{overpic}
\usepackage{enumerate}

\cvprfinalcopy 

\def\cvprPaperID{1793} 

\ifcvprfinal\fi
\begin{document}
\title{When Kernel Methods meet Feature Learning: \\ Log-Covariance Network for Action Recognition from Skeletal Data}

\author{Jacopo Cavazza$^{\pmb{1},\pmb{2}}$, Pietro Morerio$^{\pmb 1}$ and Vittorio Murino$^{ \pmb{1},\pmb{3}}$\\ \\
	$^{\pmb 1}$Pattern Analysis \& Computer Vision (PAVIS) -- Istituto Italiano di Tecnologia -- \textit{Genova, Italy} \\
	$^{\pmb 2}$Electrical, Electronics and Telecommunication Engineering and Naval Architecture Department \\ (DITEN) -- Universit\`{a} degli Studi di Genova --  \textit{Genova, Italy}\\
	$^{\pmb 3}$Computer Science Department -- Universit\`{a} di Verona --  \textit{Verona, Italy} \\
	{\tt\small \{pietro.morerio,jacopo.cavazza,vittorio.murino\}@iit.it}}

\maketitle
\thispagestyle{empty}

\begin{abstract}

Human action recognition from skeletal data is a hot research topic and important in many open domain applications of computer vision, thanks to recently introduced 3D sensors. In the literature, naive methods simply transfer off-the-shelf techniques from video to the skeletal representation. However, the current state-of-the-art is contended between to different paradigms: kernel-based methods and feature learning with (recurrent) neural networks. Both approaches show strong performances, yet they exhibit heavy, but complementary, drawbacks. Motivated by this fact, our work aims at combining together the best of the two paradigms, by proposing an approach where a shallow network is fed with a  covariance representation. Our intuition is that, as long as the dynamics is effectively modeled, there is no need for the classification network to be deep nor recurrent in order to score favorably. We validate this hypothesis in a broad experimental analysis over 6 publicly available datasets.
\end{abstract}

\begin{figure}[t!]
	\centering
	\includegraphics[width=.9\columnwidth,keepaspectratio]{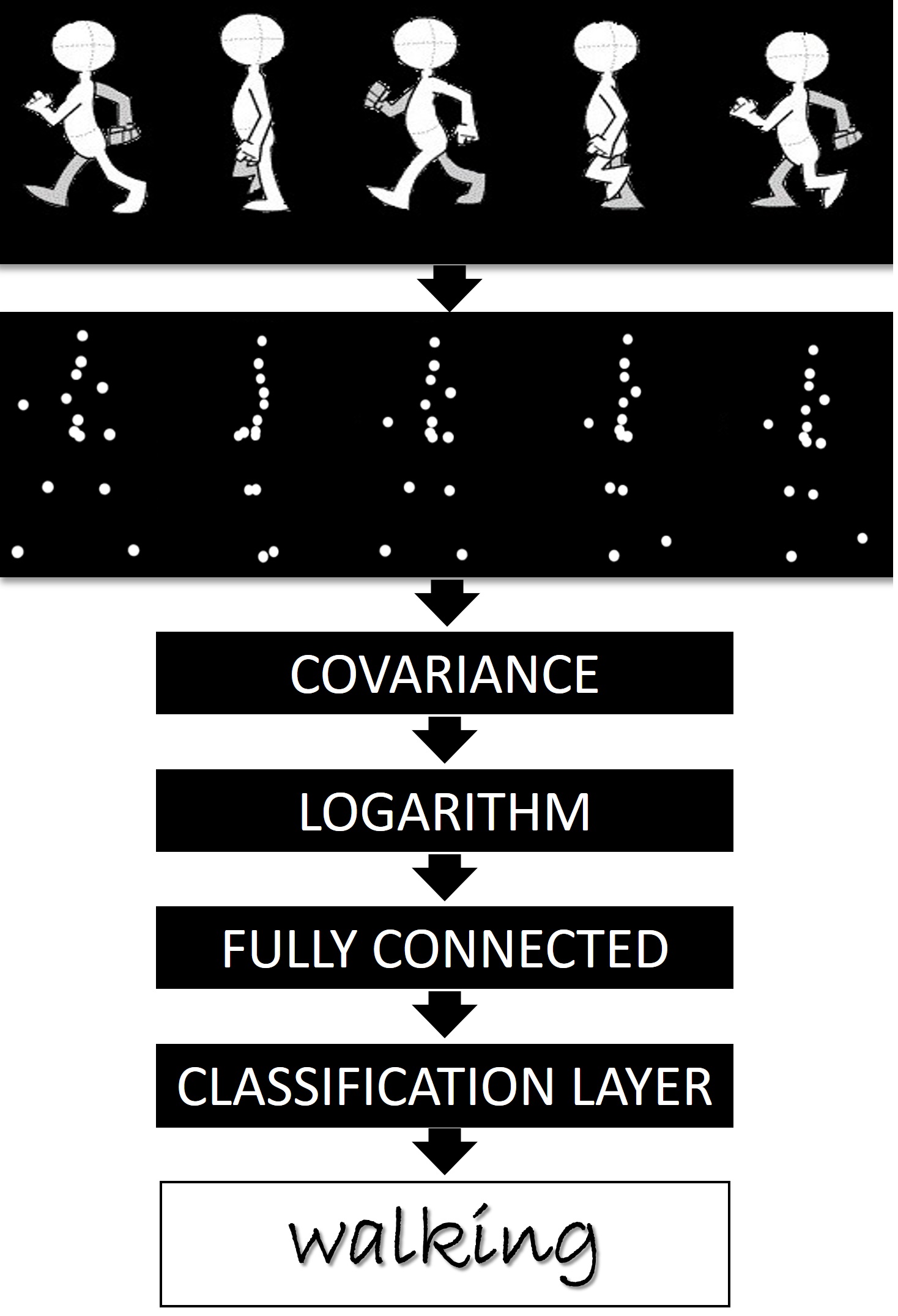}
	\caption{Overview of the proposed \textit{Log-COV-Net}. Starting from a time series of skeletal representations (top), we process the input data \eqref{eq:timeseries} with a covariance matrix \eqref{eq:cov} which is then log-projected \eqref{eq:log}. A (separately trained) fully connected layer provides the final representation for the classification stage.}
	\label{fig:pip}
\end{figure}

\section{Introduction}

Human action recognition is a paramount domain in many applicative fields, such as crowd analysis and surveillance, elderly care and autonomous driving vehicles, to name a few. 

Despite the wide interest in video-based approaches, this type of data is intrinsically affected by several issues, \eg privacy, occlusions, light variations and background noise. An effective alternative to deal with these challenges is represented by skeletal based representation. This paradigm relies on theoretical guarantees concerning motion perception. It has in fact been proved by Johansson \cite{Johansson} that the displacement of light sources located on keypoints on the humans' skeleton are enough for the visual system to recognizing the displayed action (such as walking, Fig. \ref{fig:pip}, top).

Grounding on that, the evolution of systems which can acquire the skeletal joints nowadays guarantees a reliable estimate of 3D body posture - motion capture, \eg VICON - and a cheap price - depth sensors, \eg Kinect. Additionally, replacing videos with skeletal data does not change the overall general pipeline of action classification: learning/engineering feature representation from trimmed sequences, in order to train a classifier. In practice, for a general action $\mathfrak{a}$, skeletal data is acquired in the form of the following multi-dimensional time-series.
\begin{small}
\begin{equation}\label{eq:timeseries}
\mathbf{P}_\mathfrak{a} = \begin{bmatrix}
x_1(t = 1) & x_1(t = 2) & \dots & x_1(t = T) \\
y_1(t = 1) & y_1(t = 2) & \dots & y_1(t = T) \\
z_1(t = 1) & z_1(t = 2) & \dots & z_1(t = T) \\
x_2(t = 1) & x_2(t = 2) & \dots & x_2(t = T) \\
y_2(t = 1) & y_2(t = 2) & \dots & y_2(t = T) \\
z_2(t = 1) & z_2(t = 2) & \dots & z_2(t = T) \\
\vdots & \vdots & \vdots & \vdots \\
x_J(t = 1) & x_J(t = 2) & \dots & x_J(t = T) \\
y_J(t = 1) & y_J(t = 2) & \dots & y_J(t = T) \\
z_J(t = 1) & z_J(t = 2) & \dots & z_J(t = T) 
\end{bmatrix}
\end{equation} 
\end{small}
where columns correspond to timestamps (from $t=1$ to $t=T$) and triplets of rows $[x_i(t),y_i(t),z_i(t)]^\top$ correspond to 3D spatial coordinates of the $i$-th joint, $i = 1,\dots,J$.

\subsubsection*{Related work}
In principle, if each frame in a \textit{video} was vectorized, we could gather a data matrix very similar to \eqref{eq:timeseries}. This is why, especially in the past year, many algorithms, originally devised for video-based action recognition, have been just brought to the skeletal data paradigm upon minor modifications. Among others, we can mention histogram based representations to perform temporal pooling \cite{normal,H3DJ,quads}, extraction of local spatio-temporal features from the data \cite{motraj,multibag}, also applying bag-of-words or Fisher vector approaches to aggregate the raw joint representation in a unique action descriptor \cite{quads,ela}.

However, the performance of transferring a video-based approach to skeletal data has been proved to be suboptimal with respect to more principled method which endows $\mathbf{P}_\mathfrak{a}$ with some kind of structure which can be exploited in the classification stage. We call this structured representation a \emph{kernel}. Among the many proposed kernels, we can recall the representation of each joint trajectory as a roto-translation matrix \cite{Vemulapalli:CVPR14,Vemulapalli:CVPR16,deeplie}, which leads to exploit the Lie group and Lie algebra properties of the Special Euclidean group. Alternatively, Hankel matrices \cite{Camps:ACCV14,Camps:CVPR16} have been attested to be extremely effective in the field of action recognition from skeletal data, either being paired with Hidden Markov Models \cite{Camps:ACCV14} or with a prototype-based nearest neighbor classification on the Riemannian manifold \cite{Camps:CVPR16}. Actually, in both cases of roto-translations and Hankel matrix, countermeasures (such as warping \cite{Vemulapalli:CVPR14,Vemulapalli:CVPR16}) needs to be taken against the following issue: in \eqref{eq:timeseries}, while $J$ is fixed (being an intrinsic parameter of the device used for skeleton's acquisition), $T$ is not, and can in fact changes from action to action (and even among repetition of the same action performed by the same person). Therefore, a pre-processing step (such as warping \cite{Vemulapalli:CVPR14,Vemulapalli:CVPR16}) needs to be applied in order to fix $T$ across instances, since standard methods only deal with fixed-length inputs.

In this respect, the \textit{covariance representation} is a very straightforward workaround. Formally, the covariance matrix $\mathbf{X}_\mathfrak{a}$ related to \eqref{eq:timeseries} is defined as
\begin{equation}\label{eq:cov}
\mathbf{X}_\mathfrak{a} = \dfrac{1}{T - 1} \mathbf{P}_\mathfrak{a} \left( \dfrac{1}{T}\mathbf{I}_{T} - \mathbf{1}_{T} \right) \mathbf{P}_\mathfrak{a}^\top
\end{equation}
where $\mathbf{I}_{T}$ is the identity matrix and $\mathbf{1}_{T}$ is the $T \times T$ matrix whose all entries are equal to $1$. By definition \eqref{eq:cov}, $\mathbf{X}_\mathfrak{a}$ is a $3J \times 3J$ SPD (symmetric and positive definite) matrix: we have therefore a fixed-dimensional representation, no matter which is the length $T$ of the time series \eqref{eq:timeseries}. In fact the index $t$ is saturated by the summations related to the row-by-column matrix products in \eqref{eq:cov}. 

In addition to the remarkable property of being invariant with respect to sequence length $T$, the covariance representation was proved to be an effective tool for action classification \cite{egizi,Harandi:CVPR14,Wang:ICCV15,Minh:CVPR16,Cavazza:ICPR16}. The reason for this lies in the statistical computation of second order temporal momentum of $\mathbf{P}_\mathfrak{a}$, the latter being very discriminative in recognizing human actions \cite{egizi,Wang:ICCV15,Cavazza:ICPR16}.

Recently, with the introduction of the first big dataset for action recognition with skeletal representation \cite{Shahroudy:CVPR16}, kernel methods are frequently difficult to scale up. This is due the prohibitive dimension of the training/testing Gram matrices, which compute the (kernel) pairwise similarity \emph{for every couple of instances in the dataset}. Thus, they have a quadratic cost as a function of the number of examples. Therefore, their big size make them simply intractable under a computational perspective. In order to circumvent such problem, deep learning is an alternative class of state-of-the-art approaches. In \cite{Du:CVPR15,deeplie,Shahroudy:CVPR16,Liu:ECCV16} hierarchical feature representations are learnt from the data itself, providing an end-to-end trainable encoding \& scalable classification pipeline. However, the reason for their success depends on the big number of free parameters to optimize, being the latter step complicated (the optimization is non-convex, overfitting is a real issue \cite{CV}) and computationally intense (GPU acceleration is fundamental).

In light of the dichotomy kernels vs. deep nets, many works have attempted to interconnect the two opposed paradigms. Namely, implementing kernel methods as neural networks (\eg deformable part models \cite{Girshick:CVPR15}, multiple kernel learning \cite{Rebai:NCA16}) or kernelizing existing neural network architectures (\eg convolutional kernel networks \cite{Mairal:NIPS14}, SVM neural networks \cite{Tang13}). Recently, neural networks have been tailored to be fed with structured matrices (covariance matrices \cite{deepspd} and rotation matrices \cite{deeplie}). Indeed, classical operations have been re-formulated to accommodate for the different type of input data adopted: for instance, max pooling is performed on the eigenvalues only \cite{Tang13}. 

With respect to \cite{Girshick:CVPR15,Rebai:NCA16,Mairal:NIPS14,Tang13} we notice that, in some cases, the connections between the two classes are weak in the sense that one of them impose its own formalism on the other (\eg using backpropagation in multiple kernel learning \cite{Rebai:NCA16}). Also, sometimes the connections are just a re-interpretation of existing paradigms \cite{Girshick:CVPR15}, which is theoretically interesting but not advantageous in terms of learning better models. Last, but not least, in all works \cite{Du:CVPR15,Shahroudy:CVPR16,JCNN1,Liu:ECCV16,JCNN2,deepspd,deeplie} the network adopted are deep, which actually makes the overall pipeline difficult to train, since overfitting must be controlled, the problem of local minima and saddle points must be faced, massive training data is required, as well as expensive hardware resources.

\subsubsection*{Paper Contributions} Within the previous context, our paper proposes the following main contribution.

\begin{enumerate}
	\item Within the existing literature of data-driven representation from structured input matrix \cite{Du:CVPR15,Shahroudy:CVPR16,JCNN1,Liu:ECCV16,JCNN2,deepspd,deeplie}, we propose a novel network architecture, fed by covariance representation which ultimately intertwines hand-crafted kernel methods with data-driven feature learning. 
	\item We posit that when action's kinematics is properly encoded through a kernel, there is no need to train deep architectures. Shallow networks are indeed effective.
	\item We confirm the previous intuition within a broad experimental evaluation over 6 publicly available datasets in 3D action recognition scoring a favorable performance in terms of improvement over state-of-the-art classification methods.
	\item We can recover from the scalability issue of kernel methods and mitigate the training issues of neural networks. Therefore, we achieve a strong performance and training/inference efficiency on CPU, ultimately devising an effective action recognition system for the open domain.
\end{enumerate}

\subsubsection*{Paper outline.} In Section \ref{sez:LCN}, we describe the proposed approach, called \emph{\textit{Log-COV-Net}}, for 3D action recognition from skeletal data. Section \ref{sez:e} presents a broad experimental evaluation, while Section \ref{sez:d} provides a comprehensive discussion of them. Finally, Section \ref{sez:ciao} draws conclusions, highlights limitation and profiles future work.

\section{\textit{Log-COV-Net}: Log-Covariance Network}\label{sez:LCN}

Covariance-based representations for action recognition from skeletal joints have attested a superior performance \cite{egizi,Harandi:CVPR14,Wang:ICCV15,Minh:CVPR16,Cavazza:ICPR16}. However, in order to fully exploit the structure which is induced by the covariance representation, classifiers have to be kernelized in order to fully exploit Riemannian geometry when learning decision boundaries to discriminate across different actions. Despite this being mathematically fine, some computational drawbacks arise. Indeed, training such a classifier often requires the computation of Gram matrices, whose quadratic complexity in terms of data instances makes the whole procedure intractable in the big data regime. 

On the other side, feature learning approaches via neural networks fully benefit from a gigantic amount of training examples to optimize the huge number (millions, billions) of parameters present in a deep network. At the same time, this is the main reason for the astonishing results scored by data-representations and the source of difficulty in effectively training such networks. Indeed, the optimization problem is non-convex, prone to overfitting, requiring acceleration through parallel GPU computation.

Therefore, despite the strong performance provided by either covariance-based or feature learning paradigms, each of them has its own drawbacks (scalability versus difficult training, respectively). To this end, in this work we aim at intertwining covariance-based and feature approaches in order to combine their pros and get rid of the cons. Namely, our unifying approach will achieve state-of-the-art classification, guaranteeing scalability to the big data regime and allowing easy and fast training/inference on CPU. This is possible by leveraging our intuition that, since exploiting the powerful covariance representation to encode action dynamics, there is no need for the network to be deep. In fact, shallow architectures are just enough in mining discriminative patterns for action classification. 

In the rest of this Section, we present the proposed approach called Log-Covariance Network, which is sketched in Fig. \ref{fig:pip}, and we provide and intuition for it.

\paragraph{Log-Covariance Network.} For each action instance $\mathfrak{a}$, acquired in the form of the multi-dimensional time series \eqref{eq:timeseries}, we compute a covariance matrix $\mathfrak{a}$ according to formula \eqref{eq:cov}. Then, we project $\mathbf{X}_\mathfrak{a}$ by a logarithm mapping $\log$. By exploiting the eigendecomposition
\begin{equation}
\mathbf{X}_\mathfrak{a} = \mathbf{U} \begin{bmatrix}
\lambda_1 & 0 & \dots & 0 \\
0 & \lambda_2 & \dots & 0 \\
& & \ddots & \\
0 & 0 & \dots & \lambda_{3J} \\
\end{bmatrix} \mathbf{U}^\top
\end{equation}
for $\mathbf{X}_\mathfrak{a}$, $\log \mathbf{X}_\mathfrak{a}$ is trivial to compute in as follows
\begin{equation}\label{eq:log}
\log \mathbf{X}_\mathfrak{a} = \mathbf{U} \begin{bmatrix}
\log \lambda_1 & 0 & \dots & 0 \\
0 & \log \lambda_2 & \dots & 0 \\
& & \ddots & \\
0 & 0 & \dots & \log \lambda_{3J} \\
\end{bmatrix} \mathbf{U}^\top,
\end{equation}
since all $\lambda_i$ are strictly positive. Formally, this is interpreted as a projection over the tangent space \cite{Har1,Har2,Harandi:CVPR14}, which is locally Euclidean, naturally inducing a vectorization which does not corrupt the geometry. Precisely, we define $\mathbf{v}_\mathfrak{a}$ to be the vectorization of all diagonal and lower-diagonal entries\footnote{Due to simmetry, the upper-diagonal elements are the same as the lower-diagonal ones} of $\log \mathbf{X}_\mathfrak{a}$: as similarly done in \cite{egizi,Wang:ICCV15,Cavazza:ICPR16} such intermediate representation is fully able to provide and Euclidean (vectorial) representation which keeps the powerfulness of the Riemannian encoding as SPD matrix \cite{Har1,Harandi:CVPR14,Har2}.  
Finally, the vector $\mathbf{v}_\mathfrak{a}$ is fed into a fully connected (FC) layer, followed by a sigmoid linearity, which is in turn fed into a classification layer where a hinge loss is exploited. We call our network \textit{\textit{Log-COV-Net}}.

\emph{Implementation details}. Despite all matrices $\mathbf{X}_\mathfrak{a}$ are positive definite in theory, due to numerical issues, the computed eigenvalues are not always positive: before applying the $\log$ mapping, we replace $\lambda_i$ with $\lambda_i' = \lambda_i + 10^{-4}$. 
With respect to Fig. \ref{fig:pip}, note that the ``covariance'' and ``logarithm'' layers (which implement equation \eqref{eq:cov} and \eqref{eq:log}, respectively) are parameter-free. The only parameter to be trained are the weights $\mathbf{W}$ of the fully connected layer and, of course, the ones of the final classification layer. In our experimental setup, we found that if we jointly train $\mathbf{W}$ and the classifier's parameters, we are highly sensitive to the size of the FC layer. Differently, we achieve more stability by pre-training the FC weights with a cross-entropy loss, also exploiting the powerfulness of supervision. For doing that, we use conjugate gradient descent for all experiments except the ones on NTU-RGB+D \cite{Shahroudy:CVPR16} dataset (Section \ref{sez:NTU}) where we exploit ADAM optimizer with mini-batches of 1024 elements. As a final step, we separately train the hinge-loss classification layer (using libLINEAR \cite{liblinear}). 

\section{Experiments}\label{sez:e}

We present here the classification accuracies registered by the proposed \textit{Log-COV-Net} in a broad comparison with several state-of-the-art methods on a plethora of benchmark datasets. Namely, we evaluated on  MSR-Action3D \cite{Action3D}, MSR-Action-Pairs (MSR-$pairs$) \cite{MSRPairs},  Gaming-3D (G3D) \cite{G3D}, Florence3D \cite{Florence3D}, UTKinect \cite{UTKinect},  MSRC-Kinect12 \cite{MSRC}, HDM-05 \cite{HDM-05} and NTU RGB$+$D \cite{Shahroudy:CVPR16}. In all cases, we followed the respective recommended training/testing protocols.

As a common preprocessing steps, we compute the relative difference of each joint' triplets $[x_i(t),y_i(t),z_i(t)]^\top$ with the position of the root joint $[x_{\rm root}(t),y_{\rm root}(t),z_{\rm root}(t)]^\top$ for any $t$. Typically the hip center is adopted as the root. This reduces the actual dimension of $\mathbf{X}_\mathfrak{a}$ to $3(J-1) \times 3(J-1)$. Also, the size of the FC layer was cross-validated through grid search within 8, 16, 36, 64, 128, 256 and 512. 

\subsection{MSR-Action 3D} The dataset consists of 20 actions
from 10 different subjects, and is collected with a depth sensor. Each subject performed every action twice or more (total 557 sequences). The 3D locations of 20 joints are provided with the dataset. This is a challenging dataset because many of the actions are highly similar to each other.

\paragraph{Comparative analysis} We benchmarked the proposed (\textit{Log-COV-Net}) against the Hankel-based approaches \cite{Camps:ACCV14,Camps:CVPR16}, used in tandem with either a Hidden Markov Model (HMM) or a Riemannian-nearest neighbors classifier with prototypes (Hankel-NN-proto). Also, we compared against the tensor representation provided by \cite{ECCV16} in using Sequence and Dynamics Compatibility Kernels (SCK+DCK). For covariance based approaches, we also included Kernelized-COV, the kernelization proposed by \cite{Cavazza:ICPR16}. We considered the kernel networks of \cite{Wang:ICCV15} where trial-specific Gram matrix are fed in a second-level kernel responsible for ultimate classification. Finally, we included the LSTM-based approach of \cite{Liu:ECCV16} where a graph represents skeleton's geometry. 

Our experimental findings on MSR-Action-3D are reported in Table \ref{tab:MSRA}.

\begin{table}[h!]
	\centering
\begin{tabular}{|rc|}
	\hline
	Hankel-HMM \cite{Camps:ACCV14} & 89.0\% \\
	SCK + DCK \cite{ECCV16} & 94.0\% \\
	Hankel-NN-proto \cite{Camps:CVPR16} & 94.7\% \\
	graph-joint-LSTM \cite{Liu:ECCV16} & 94.8\% \\
	Kernelized-COV \cite{Cavazza:ICPR16} & 96.8\% \\ 
	Ker-RP-RBF \cite{Wang:ICCV15} & 96.9\% \\ \hline \hline
	\textit{Log-COV-Net} (proposed) & \textbf{\textit{97.4\%}} \\ \hline
\end{tabular}
\caption{Evaluation on MSR-Action-3D using the protocol of \cite{Action3D}.}
\label{tab:MSRA}
\end{table}

%
%
%
%

\subsection{Gaming 3D}

The dataset includes 20 different gaming actions like \textsf{golf swing}, \textsf{tennis serve} or \textsf{bowling}. 10 subjects were involved in the acquisition, each of them performing each action three or more times for a total of 663 action sequences, represented by the displacement in time of 20 joints.

\paragraph{Comparative analysis} We benchmarked with \cite{RBMHMM} which combined Restricted Boltzmann machines and Hidden Markov Models. Also, we included several Lie geometry-based methods to encode roto-translations: the shallow approaches Lie Group \cite{Vemulapalli:CVPR14} and Lie Algebra \cite{Vemulapalli:CVPR16} and the intertwined Lie/deep method \cite{deeplie}. For the latter, we selected LieNet-3Block which is the best performing architecture within the ones proposed in \cite{deeplie}.

Results are reported in Table \ref{tab:G3D}.

\begin{table}[h!]
	\centering
	\begin{tabular}{|rc|}
	\hline
	RBM + HMM \cite{RBMHMM} & 86.4\% \\
	LieNet-3Blocks \cite{deeplie} & 89.1\% \\
	Lie Algebra \cite{Vemulapalli:CVPR16} & 90.9\% \\
	Lie Group \cite{Vemulapalli:CVPR14} & 91.1\% \\ \hline \hline
	\textit{Log-COV-Net} (proposed) & \textbf{\textit{93.0\%}} \\ \hline
\end{tabular}\vspace{5pt}
\caption{Evaluation on Gaming 3D using the protocol of \cite{Vemulapalli:CVPR16}.}
\label{tab:G3D}
\end{table}

\subsection{UT Kinect}

This dataset was captured using a stationary Kinect sensor, the 3D locations of 20 joints are provided. 10 different subjects perform 10 different actions (twice each). This is a challenging dataset due to variations in the view point and high intra-class variations.

\paragraph{Comparative analysis} Our proposed \textit{Log-COV-Net} is compared against Lie Group representation \cite{Vemulapalli:CVPR14}, HMM fed with Hankel matrices \cite{Camps:ACCV14}, PCA on manifold \cite{ela} and the LSTM with graph-based encoding of human skeleton \cite{Liu:ECCV16}. In addition, we also compared with the reformulation of Histogramd of Oriented Gradients (HOG) features for joints \cite{H3DJ} and the aggregation of local spatio-temporal features extracted from raw data \cite{randforest}.

Results are reported in Table \ref{tab:UT}.

\begin{table}[h!]
	\centering
	\begin{tabular}{|rc|}
	\hline
	Histograms of 3D joints \cite{H3DJ} & 90.9\% \\
	Spatio-temporal local features \cite{randforest} & 87.9\% \\ 
	Lie Group \cite{Vemulapalli:CVPR14} & 97.1\% \\ 
	Hankel-HMM \cite{Camps:ACCV14} & 86.8\% \\
	Manifold PCA \cite{ela} & 94.9\% \\
	graph-joint-LSTM \cite{Liu:ECCV16} & 97.0\% \\ \hline \hline
	\textit{\textit{Log-COV-Net} } (proposed) & \textit{\textbf{98.3\%}} \\ \hline
\end{tabular}\vspace{5pt}
	\caption{Evaluation on UT Kinect using the protocol of \cite{Vemulapalli:CVPR14}.}
\label{tab:UT}
\end{table}

\subsection{MSRC-Kinect 12}

An acquisition of six hours and 40 minutes involves 30 people performing 12 gestures. In total, 6,244 gesture instances. The motion files contain Kinect estimated trajectories of 20 joints. 

\paragraph{Comparative analysis} Several covariance-based approaches are compared against the proposed \textit{Log-COV-Net}. Precisely, we considered the Bregman divergences for the infinite dimensional operators \cite{Harandi:CVPR14},  the temporal pyramid of covariance descriptors \cite{egizi} and the kernelization recently provided by \cite{Cavazza:ICPR16}. Also, we included Ker-RP-POL and Ker-RP-RBF, the two kernel networks of \cite{Wang:ICCV15}.

Results are reported in Table \ref{tab:MS}.

\begin{table}[h!]
	\centering
	\begin{tabular}{|rc|}
	\hline
	Bregman-div \cite{Harandi:CVPR14} & 89.9\%   \\
	Ker-RP-POL \cite{Wang:ICCV15} & 90.5\%  \\
	Ker-RP-RBF \cite{Wang:ICCV15} & 92.3\%  \\
	Pyramid of COV \cite{egizi} & 93.6\% \\
	Ker-COV \cite{Cavazza:ICPR16} & 95.0\% \\ \hline \hline
	\textit{Log-COV-Net} (proposed) & \textbf{\textit{98.5\%}} \\ \hline
\end{tabular}\vspace{5pt}
	\caption{Evaluation on MSRC Kinect 12 using the protocol of \cite{egizi}.}
\label{tab:MS}
\end{table}

\subsection{HDM-05}

This dataset contains more than tree hours of systematically recorded and well-documented motion capture data using a 240Hz VICON system to acquire the gestures of 5 non-professional actors via 31 markers. Motion clips have been manually cut out and annotated into roughly 100 different motion classes: on average, 10-50 realizations per class are available. In order to be consistent with the literature, we both replicate the 14 classes evaluation \cite{Wang:ICCV15,Cavazza:ICPR16} and report the results on the whole dataset.

\begin{table}[h!]
	\centering
	\begin{tabular}{|rcc|}
		\hline
		& \textit{14-classes} & \textit{all-classes} \\ \hline\hline
		sparse-D-SPD \cite{Har1} & 76.1\% & N.A. \\
		COV-discriminative \cite{COVdiscr} & 79.8\% & N.A. \\
		SPD-dim-red \cite{Har2} & 81.9\% & 40.0\% \\
		Bregman-divergence \cite{Harandi:CVPR14}  & 82.5\% & N.A. \\
		Hankel-NN-proto \cite{Camps:CVPR16} & 86.3\% & N.A. \\
		Region-COV \cite{Tuzel} & 91.5\% & 58.9\% \\
		Ker-RP-POL \cite{Wang:ICCV15} & 93.6\% & 64.3\% \\
		Ker-RP-RBF \cite{Wang:ICCV15} & 96.8\% & 66.2\% \\ \hline \hline
		\textit{\textit{Log-COV-Net} } (proposed) & \textit{\textbf{99.1\%}} & \textit{\textbf{72.0\%}} \\ \hline
	\end{tabular}\vspace{5pt}
	\caption{Evaluation on HDM-05 using the two protocols of \cite{Wang:ICCV15}.}
	\label{tab:05}
\end{table}

\paragraph{Comparative analysis} In a broad experimental validation of \textit{Log-COV-Net}, we considered the sparse coding with dictionary learning for SPD matrices of \cite{Har1}, the covariance discriminative learning framework of \cite{COVdiscr}, and the dimensionality reduction technique of \cite{Har2} for SPD matrices. In addition to Bregman-divergence of the infinite covariance representation \cite{Harandi:CVPR14} and the fast region covariance descriptor of \cite{Tuzel}, we included the trial-specific encoding of an action with a Gram matrix \cite{Wang:ICCV15} reporting both Ker-RP-POL and Ker-RP-RBF (polynomial vs. Gaussian RBF kernel). We also compared against \cite{Camps:CVPR16}.

Resultsare reported in Table \ref{tab:05}, adding the performance of our \textit{Log-COV-Net} to the published results of \cite[Table 4.]{Wang:ICCV15}.

\subsection{NTU RGB+D}\label{sez:NTU}

This huge dataset contains 60 different action classes including daily, mutual, and health-related actions. 40 subjects where involved in the acquisition, for a total number of about 60K instances, captured from 3 different views. According the suggested experimental protocols \cite{Shahroudy:CVPR16}, we performed a cross validation by testing the model on either different subjects or view with respect to the ones used in training.

\paragraph{Comparative analysis} We compared the proposed Log-COV-Net on the NTU-RGB+D dataset. We benchmarked the approaches \cite{normal,Hnormal} which rely on normal vector computations, either with a temporal pooling of 3D normals \cite{Hnormal} or with modeling of 3D+time spatio-temporal coordinates as a whole \cite{normal}. We included the generalization of HOG for skeletal joints \cite{H3DJ}, also considering the aggregation of raw joints data by means a Gaussian mixture model and Fisher Vectors extraction. We additionally reported the performance of Lie geometry representation by either directly employing Lie Group structure \cite{Vemulapalli:CVPR14} or the Riemannian-training of a neural network \cite{deeplie}.  

\begin{table}[h!]
	\centering
	\begin{tabular}{|rcc|}
		\hline
		& {\footnotesize \textit{cross-subject}} & {\footnotesize \textit{cross-view}} \\ \hline\hline
		Histogram of 3D Normals \cite{Hnormal} & 30.6\% & 7.3\% \\
		4D Normal Vectors \cite{normal} & 31.8\% & 13.6\% \\
		Histograms of 3D joints \cite{H3DJ} & 32.4\% & 22.3\% \\
		Fisher Vectors \cite{quads} & 38.6\% & 41.4\% \\
		Lie Group \cite{Vemulapalli:CVPR14} & 50.1\% & 52.8\% \\  
		HB-RNN \cite{Du:CVPR15} & 56.3\% & 64.0\% \\
		joint-RNN \cite{Shahroudy:CVPR16}  & 59.1\% & 64.1\% \\
		LieNet-3Blocks \cite{deeplie} & 61.4\% & 67.0\% \\
		joint-LSTM \cite{Shahroudy:CVPR16} & 60.7\% & 67.3\% \\
		graph-joint-LSTM \cite{Liu:ECCV16} & \textbf{69.2\%} & \textbf{77.7\%} \\ \hline \hline
		\textit{\textit{Log-COV-Net} } (proposed) & \textit{60.9\%} & \textit{63.4\%} \\ \hline
	\end{tabular}\vspace{5pt}
	\caption{Evaluation on NTU RGB+D with two protocols of \cite{Shahroudy:CVPR16}.}
	\label{tab:NTU}
\end{table}

Actually, the release of NTU-RGB+D in 2016 promoted a big boost in training deep architectures for action recognition from skeletal joints. Within the most effective neural network approaches, Recurrent Neural Network (RNN) play a pivotal role. Indeed, \cite{Shahroudy:CVPR16} trained a RNN by directly feeding raw skeletal data. \cite{Du:CVPR15} performed a hierarchical decomposition of human skeleton into arms-legs-torso, modeling each of them with a network and fusing all scores in a bottom-up fashion. Long-short term memory units boosts RNN: \cite{Shahroudy:CVPR16} used them from raw joints and \cite{Liu:ECCV16} directly encoded the skeletal geometry by means of a graph. 

Results are reported in Table \ref{tab:NTU}


\begin{table*}[t!]
	\centering
	{
	\renewcommand{\arraystretch}{1.2}
	\begin{tabular}{|c|c|c|c|}
		\hline
		MSR-Action 3D  & Gaming 3D & UT Kinect & HDM-05, \textit{14-classes} \\\hline\hline
		$N \sim 10^2$  & $N \sim 10^2$ &  $N \sim 10^2$ & $N \sim 10^2$ \\
		$\iota =$ \textbf{+0.5\%} & $\iota =$  \textbf{+1.9\%} & $\iota =$  \textbf{+1.3\%} & $\iota =$ \textbf{+2.3\%} \\\hline\hline
		MSRC-Kinect12 & HDM-05, \textit{all-classes} & NTU-RGB+D, \textit{\footnotesize cross-subject} & NTU-RGB+D, \textit{\footnotesize cross-view} \\ \hline\hline		
		$N \sim 10^3$ & $N \sim 10^3$ & $N \sim 10^4$ & $N \sim 10^4$ \\
		$\iota =$ \textbf{+3.5\%} & $\iota =$  \textbf{+5.8\%} & $\iota =$ -8.3\% & $\iota =$  -14.3\% \\ \hline
	\end{tabular}
	}	
	\vspace{5pt}
	\caption{Comprehensive evaluation of the proposed approach, measuring the (positive or negative) improvement $\iota$ of the proposed \textit{Log-COV-Net} with respect to the best among the reported methods. For any dataset, we also provide $N$, that is the order of magnitude of the available instances.}
	\label{tab:glob}
\end{table*}

\section{Discussion}\label{sez:d}

In this Section we analyze all the results scored on the datasets we consider (Tables \ref{tab:MSRA}, 
\ref{tab:G3D}, 
\ref{tab:UT}, \ref{tab:MS}, \ref{tab:05} and \ref{tab:NTU}). The discussion is carried on according to the cardinality of the training sets, as provided in Table \ref{tab:glob}.

\begin{description}
	\item[$N \sim 10^2 \; - \;$] In the small data regime, the amount of examples available does not allow to fully benefit from the learning from data paradigm. Nevertheless, our proposed \textit{Log-COV-Net} is performing on par with respect to the best method reported hereby (0.2\% negative gap on Florence 3D), while improving all baselines in all other cases with about 1\% on average.
	\item[$N \sim 10^3 \; - \;$] Increasing by factor 10, we achieve a medium data regime which attested to be the ideal setting for our proposed shallow network. In such a case we register outstanding improvements of \textit{Log-COV-Net} over the state-of-the-art: +3.5\% on MSRC-Kinect 12 and +5.8\% in the all-class case for HDM-05. This is a strong empirical evidence that the combination of a powerful temporal encoding (through covariance) allows a shallow net to achieve a top performance.
	\item[$N \sim 10^4 \; - \;$] When moving to the big data regime, we have lots of training data and the relatively small number of free parameters in \textit{Log-COV-Net} does not fully capture all available discriminants. Indeed, \textit{Log-COV-Net} is quite gapped by LSTM architectures\footnote{Note that, on the small data regime, our \textit{Log-COV-Net} is superior to this architectures: \eg, Tab. \ref{tab:UT}, +1.3\% on graph-joint-LSTM \cite{Liu:ECCV16}.} \cite{Shahroudy:CVPR16,Liu:ECCV16}. In spite of that, we can nevertheless see that all hand-crafted approaches \cite{normal,Hnormal,H3DJ,quads,Vemulapalli:CVPR14} are greatly outperformed in performance, but also scoring on par with respect to alternative deep architectures (e.g. \cite{deeplie} on \textit{cross-subject} protocol or the hierarchical RNN on the \textit{cross-view}). Again, an effective kinematic encoding allows a shallow net to score a strong overall performance. 
\end{description}

One further reason for our method to be appealing for open domain action recognition systems is the computational efficiency. Indeed, we adopt a very different perspective from main approaches in the literature. Indeed, we avoid dictionary-based or general pooling aggregation techniques (which slow training) or expensive computational pre-processing such as temporal warping of sequences in order to achieve a fixed temporal length \cite{Vemulapalli:CVPR14,Vemulapalli:CVPR16}. Additionally, we can simply run our training/inference stage on CPU: 20-30 minutes for training \textit{Log-COV-Net} on NTU RGB+D \cite{Shahroudy:CVPR16}, with almost realtime inference. If compared to \cite{Shahroudy:CVPR16,Liu:ECCV16,deeplie,Du:CVPR15}, the training time in this case is much longer even if using GPU acceleration. Last, but not least, we achieve a quite compact feature representation (the size of the FC is 512 at maximum), which is much much smaller with respect to other approaches, such as \cite{ECCV16} or \cite{Vemulapalli:CVPR14}. 
	
\emph{Thanks to such a compact representation, paired with an extreme training/computational efficiency, we bring strong evidence of the effectiveness of the proposed \textit{Log-COV-Net}.}

\section{Conclusion, Limitations and Future Work}\label{sez:ciao}

In this work we intertwine kernel methods and feature learning by proposing \textit{Log-COV-Net}, a shallow network fed with log-projected covariance representation of skeletal joints data for action recognition. 

We empirically prove that, after a powerful structured encoding of action dynamics, there is no actual need to train deep networks for achieving state-of-the-art performance, being a shallow configuration simply enough. Such finding results in an extremely optimized pipeline which can be trained on CPU very fast, performing action classification very efficiently and also relying on a much more compact data representation. 

Despite the overall performance is good on the small data regime (hundreds of examples) and remarkable on thousands of instances, one additional order of magnitude  ($10^4$) makes our \textit{Log-COV-Net} suffer with respect to the more elaborated LSTM (which are yet more difficult to train than our network).

Therefore, as a future work, we intend to fill this gap, still preserving compactness of representation and efficiency for training/inference on CPU.

{\small
	\bibliographystyle{ieee}
	\bibliography{egbib}
}
	
\end{document}